\documentclass[journal]{IEEEtran}

\usepackage{cite}
\usepackage{amsmath,amssymb,amsfonts}
\usepackage{algorithmic}
\usepackage{graphicx}
\usepackage{textcomp}
\usepackage{float}
\usepackage{titlesec}
\usepackage{xcolor}
\usepackage{multicol} 
\usepackage[utf8]{inputenc}  
\usepackage{caption}
\usepackage{booktabs}
\usepackage[colorlinks=true, linkcolor=blue, citecolor=blue, urlcolor=blue]{hyperref}
\usepackage[noabbrev,capitalize]{cleveref}
\titleformat{\section}[block]{\sffamily\small}{\thesection}{1em}{} 
\titlespacing*{\section}{5pt}{5pt}{5pt}
\begin{document}

\title{ LWMSCNN-SE: A Lightweight Multi-Scale Network for Efficient Maize Disease Classification on Edge Devices

}

\author{
    \IEEEauthorblockN{1\textsuperscript{st} Fikadu Weloday}
    , \IEEEauthorblockN{2\textsuperscript{nd} Jianmei Su}
    \\
    \IEEEauthorblockA{\textit{School of Computer Science and Technology,Southwest University of Science and Technology, Mianyang,China}}
    \IEEEauthorblockA{deyu@mails.swust.edu.cn}
}

\maketitle
\begin{abstract}
Maize disease classification plays a vital role in mitigating yield losses and ensuring food security. However, the deployment of traditional disease detection models in resource-constrained environments, such as those using smartphones and drones, faces challenges due to high computational costs. To address these challenges, we propose LWMSCNN-SE, a lightweight convolutional neural network (CNN) that integrates multi-scale feature extraction, depthwise separable convolutions, and squeeze-and-Excitation (SE) attention mechanisms. This novel combination enables the model to achieve 96.63\% classification accuracy with only 241,348 parameters and 0.666 GFLOPs, making it suitable for real-time deployment in field applications. Our approach addresses the accuracy--efficiency trade-off by delivering high accuracy while maintaining low computational costs, demonstrating its potential for efficient maize disease diagnosis on edge devices in precision farming systems.
\end{abstract}

\begin{IEEEkeywords}
lightweight CNN, multi-scale feature extraction, attention mechanism, plant pathology
\end{IEEEkeywords}

\section{Introduction}
Maize is one of the most widely cultivated cereal crops worldwide, playing a crucial role in global food security. However, maize productivity is severely affected by leaf diseases such as Cercospora leaf spot, common rust, and northern leaf blight, which can lead to significant yield losses if not detected and managed at an early stage \cite{Sibiya2019,subbarayudu2025,nsibo2024}. It is estimated that maize diseases contribute to 22–23\% annual yield losses globally, particularly in developing regions where access to expert agronomic support is limited\cite{nsibo2024,savary2019}.

Traditional disease diagnosis relies primarily on manual visual inspection by trained experts. While effective, this approach is time-consuming, subjective, and impractical for large-scale monitoring, especially in remote and resource-limited agricultural environments\cite{sharma2025}. These limitations highlight the need for automated, accurate, and scalable disease detection systems that can support farmers in real-world conditions \cite{dulla2025}.

Recent advances in deep learning, particularly convolutional neural networks (CNNs), have demonstrated remarkable success in plant disease classification and recognition tasks \cite{sladojevic2016,lu2021,abade2020}. State-of-the-art architectures such as VGG, ResNet, DenseNet, and EfficientNet achieve high accuracy by leveraging deep and complex network designs \cite{shafik2024,dixit2025}. However, these models typically involve millions of parameters and high computational complexity, making them unsuitable for deployment on mobile devices, unmanned aerial vehicles (UAVs), and edge-based agricultural systems \cite{khan2023}.

To address these challenges, lightweight CNN architectures such as MobileNet and ShuffleNet employ depthwise separable convolutions and efficient design principles to reduce computational cost . Although these models improve efficiency, aggressive parameter reduction can compromise feature representation capability, leading to reduced accuracy when applied to complex plant disease patterns under varying environmental conditions\cite{jiang2022,xiang2021}.

This paper proposes LWMSCNN-SE, a lightweight multi-scale CNN designed for efficient maize disease classification. Our approach integrates multi-scale feature extraction, depthwise separable convolutions, and the SE attention mechanism, enabling the model to effectively capture both local and contextual features. The model achieves competitive classification accuracy (96.63\%) while maintaining a minimal parameter count (241,348 parameters) and low computational complexity (0.666 GFLOPs), making it suitable for deployment in resource-constrained environments such as smartphones, drones, and IoT sensors.

The primary contributions of this work are:
\begin{enumerate}
    \item A lightweight multi-scale CNN architecture that effectively captures both local and contextual features using residual depthwise convolutions.
    \item Integration of squeeze-and-excitation attention to enhance channel-wise feature representation with minimal additional cost.
    \item A highly efficient model design with only 241k parameters and 0.666 GFLOPs, suitable for edge and mobile deployment.
    \item Extensive experimental evaluation demonstrating high accuracy and favorable efficiency–accuracy trade-off for maize disease classification.
\end{enumerate}

\section{Related Work}
        
Automated detection of maize leaf diseases is critical for mitigating yield losses and safeguarding global food security, particularly in regions where maize is a staple crop and a primary source of income\cite{Haque2022,Joseph2024}. In many smallholder and resource‑constrained farming systems, disease diagnosis still relies on manual field scouting by agronomists or extension workers, which, although often accurate, is inherently slow, labor-intensive, and subjective, and cannot feasibly cover large production areas at high temporal frequency. Farmers frequently lack timely access to expert advice, leading to delayed or incorrect treatment decisions, inefficient use of pesticides, and ultimately substantial yield and economic losses \cite{Shi2023}. These practical constraints have motivated the development of automated, vision-based decision support systems that can provide rapid, consistent, and scalable disease assessment directly in the field using commodity hardware such as smartphones, low-cost cameras, drones, and IoT sensor nodes\cite{FKhan2023}.

Recent advances in computer vision and deep learning have transformed plant disease recognition, with convolutional neural networks (CNNs) emerging as the widely accepted standard for image-based diagnosis. Deep CNN architectures trained on large, curated datasets have achieved highly accurate plant disease and maize‑specific classifiers, often reporting accuracies above 95–99\% under controlled conditions. Canonical high‑capacity models such as VGG16, ResNet‑50, and DenseNet exploit very deep hierarchies of convolutional layers, residual connections, and dense feature reuse to learn rich, discriminative feature representations capable of capturing complex texture, color, and shape patterns associated with different disease stages and severities \cite{Andrew2022,Chowdhury2021}. However, these high‑performing architectures typically contain tens of millions of parameters and require hundreds of megaflops to gigaflops of computation per inference, resulting in substantial memory, bandwidth, and energy demands that are impractical for deployment on mobile and embedded platforms commonly used in smart agriculture scenarios. Consequently, there is a growing accuracy–efficiency gap between models designed for data‑center GPUs and the stringent resource budgets of edge devices in real farms \cite{Li2024,Li2023}.
        
\indent To address efficiency concerns, lightweight CNN architectures including MobileNet, ShuffleNet, and EfficientNet have been proposed. These models significantly reduce parameter count and computational complexity using depthwise separable convolutions and optimized network design\cite{Howard2017,Zhang2018,Tan2019,Nguyen2020}. Nevertheless, their general-purpose nature may limit performance in plant disease classification, where visual symptoms can be subtle and highly variable \cite{jiang2022}.
        
\indent To overcome this, recent work integrates Squeeze and Excitation (SE) channel attention into lightweight backbones (e.g., EfficientNet‑SE, Tiny‑MobileNet‑SE, depthwise‑SE CNNs), showing that SE can selectively amplify informative channels with minimal extra cost, significantly boosting accuracy on crop pest and disease datasets while keeping model size and FLOPs suitable for mobile deployment. For example, EfficientNet‑SE and Tiny‑MobileNet‑SE achieve around 91–98\% accuracy with sub‑MB models on crop leaf and pest datasets 14, while depthwise CNNs with SE and residual links reach ~98\% accuracy with strong efficiency 16. Parallel efforts in plant disease recognition adopt multi‑scale feature fusion to better capture lesions of varying size and context, using parallel or serial convolutional paths and fusion strategies to enhance discrimination on wheat, maize, tea, and generic plant leaf disease datasets. Multi‑scale feature fusion shuffle networks and YOLO‑style maize detectors, for instance, employ inflated or variable‑kernel convolutions plus pyramid pooling to improve identification in complex field scenes while remaining relatively light. Yet, general multi‑scale CNNs like DeepLab‑style frameworks and heavy hybrid CNN–Transformer models often introduce substantial depth and large kernels, limiting suitability for real‑time edge deployment despite excellent segmentation or classification performance \cite{Wang2025,AlGaashani2025,Shandilya2025}. 
            
This paper introduces LWMSCNN-SE, a lightweight multi-scale CNN designed to bridge the accuracy-efficiency gap for maize disease classification. While prior work has primarily pursued two separate paths—developing general-purpose lightweight backbones (e.g., MobileNet) or enhancing accuracy via attention or multi-scale modules in larger networks—these approaches often fall short for agricultural edge deployment. General-purpose models may lack the specialized representational capacity for subtle disease patterns, while adding complex modules to standard CNNs typically inflates computational cost.

Our approach addresses this gap through a synergistic co-design of efficiency and discriminative power. Specifically, LWMSCNN-SE integrates multi-scale feature extraction and a Squeeze-and-Excitation attention mechanism directly within a depthwise separable convolutional framework. This novel integration ensures the model natively captures both local lesions and broader contextual features—critical for accurate disease identification—while maintaining an extremely low parameter and FLOP budget. Consequently, LWMSCNN-SE achieves high accuracy without compromising suitability for real-time inference on resource-constrained edge devices like smartphones, drones, and IoT sensors in precision agriculture.
         
\section{Proposed Methodology}
\begin{figure}[ht]
    \centering
    \includegraphics[width=1\textwidth]{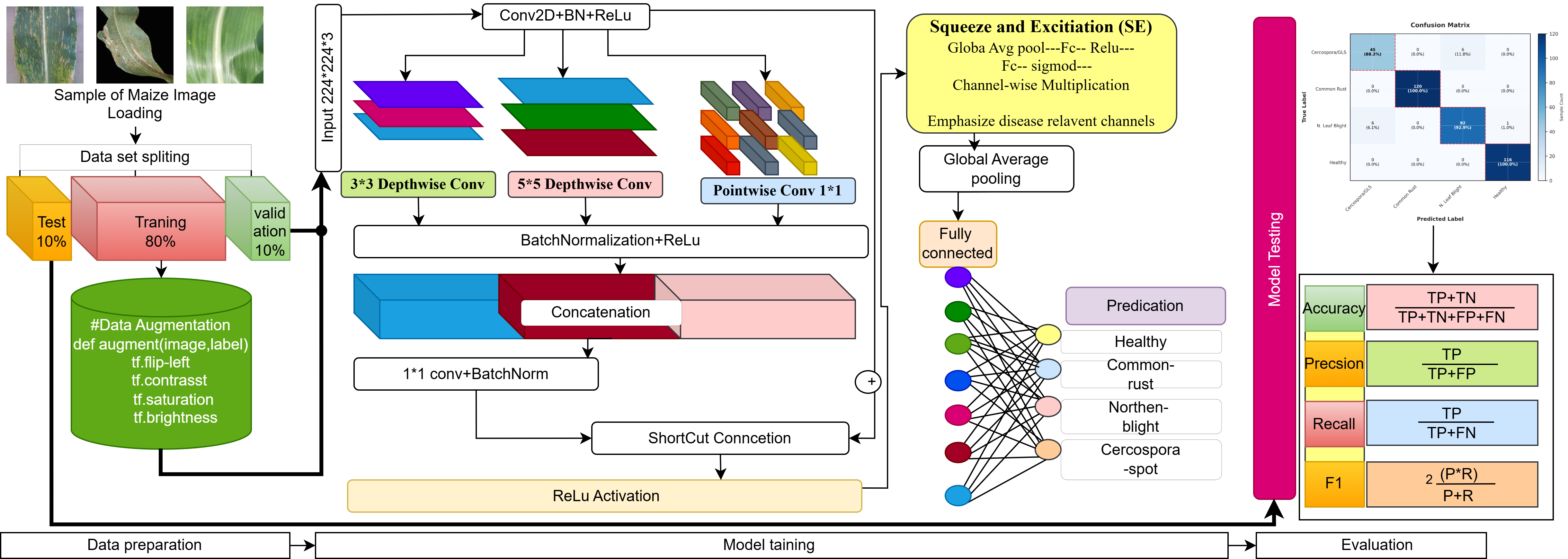} 
    \caption{End-to-end workflow of the proposed LWMSCNN-SE model for maize disease classification.}
    \label{fig:Figure1}
 \end{figure}

The proposed methodology, LWMSCNN‑SE, is a lightweight multi‑scale convolutional neural network enhanced with Squeeze‑and‑Excitation (SE) attention, designed for efficient plant disease classification in resource‑constrained agricultural environments. The model is evaluated on a maize disease dataset of 3,852 images across four disease classes. An 80:10:10 stratified split is used to partition the data into training, validation, and test sets, ensuring a representative distribution of classes in each subset. Input images are resized to 224×224 pixels, normalized to the range, and augmented with random horizontal/vertical flips and color jittering (brightness, contrast, saturation, hue) to improve generalization. The overall workflow of the proposed approach is illustrated in Figure~\ref{fig:Figure1}, which depicts the complete pipeline from input preprocessing to final classification. The architecture employs residual multi‑scale blocks that use parallel $3\times3$ and $5\times5$ depthwise convolutions to capture multi‑scale spatial patterns, combined with depthwise separable convolutions to reduce computational cost. Channel‑wise feature responses are adaptively recalibrated via SE blocks, enhancing the model’s discriminative capability. The network is trained end‑to‑end using the Adam optimizer with sparse categorical cross‑entropy loss, incorporating early stopping and model checkpointing to prevent overfitting. Performance is evaluated on the held‑out test set using accuracy, classification report, and confusion matrix.

 \newpage
 \onecolumn
 \begin{minipage}{\textwidth}
 \setlength{\parindent}{15pt}
    \begin{multicols}{2} %
         \subsection{Data Description and Preprocessing}
      The dataset used in this study consists of 3,852 RGB images from a publicly available maize disease dataset, categorized into four classes: Gray Leaf Spot (513 images), Common Rust (1,192 images), Healthy Leaves (1,162 images), and Northern Leaf Blight (985 images). For effective input into the convolutional neural network (CNN), several preprocessing steps were applied. Initially, all images were resized to a uniform dimension of 224×224 pixels. This resizing ensures compatibility with the model’s input size while also reducing computational complexity during training. The aspect ratio of the original images was maintained to avoid distortion, thereby preserving critical visual features necessary for accurate disease classification. Pixel normalization was then performed, scaling the pixel values to the range [0, 1] by dividing each pixel by 255. This normalization standardizes the input data, mitigating the impact of large numerical values and contributing to the stability and efficiency of the gradient descent optimization process, which in turn accelerates the model’s convergence. To further improve model performance and prevent class imbalance, an 80:10:10 stratified data split was employed, ensuring that each subset (training, validation, and test) maintained the same class distribution as the original dataset. This stratified splitting minimizes the risk of data leakage and ensures that the model’s evaluation remains reliable and unbiased.
\subsection{Data Augmentation}
To enhance the model’s generalization capability and address the limited dataset size, an extensive data augmentation pipeline was implemented during the training phase. This pipeline aimed to introduce variability in the input images, improving the model’s robustness to real-world conditions. The augmentation techniques employed are as follows:
         \begin{itemize}
             \item Random horizontal and vertical flips to simulate varying capture orientations.
             \item Brightness adjustments (\(\pm 20\%\)) to mimic different lighting conditions.
             \item Contrast variations (\(0.8 - 1.2 \times\)) to enhance feature visibility.
             
        \columnbreak

        \item Saturation modifications (\(0.8 - 1.2 \times\)) to account for color intensity differences.     
             \item Hue variations (\(\pm 0.05\)) to accommodate natural color variations in field conditions.
         \end{itemize}

         \noindent These augmentation techniques effectively expanded the dataset by introducing additional variability in the input images. This approach helps reduce the risk of overfitting and enhances the model’s ability to generalize to unseen data. By mimicking the natural diversity found in real-world agricultural settings, the augmented dataset improved the model's performance and robustness in practical deployment scenarios.

         \subsection{ Overall Architecture}
         The detailed model architecture is illustrated in Figure~\ref{fig:Figure2}.
        The proposed network follows a hierarchical feature extraction strategy, starting with an initial convolutional stem and followed by a sequence of residual multi-scale blocks. Depthwise convolutions are utilized to reduce the number of parameters and computational cost, while squeeze-and-excitation blocks are integrated to enhance the learning of discriminative features by adaptively recalibrating channel-wise feature responses.
       \begin{figure}[H]
           \centering
           \includegraphics[width=0.5\textwidth]{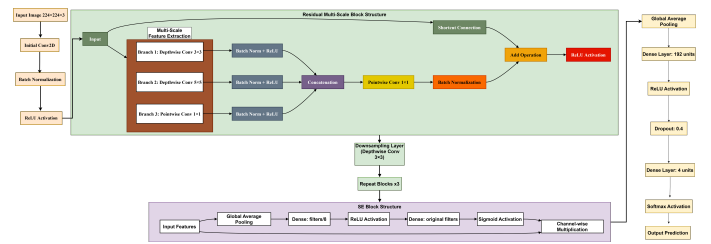}
           \caption{The architecture of the proposed LWMSCNN-SE model.}
          \label{fig:Figure2}
         \end{figure}
         
        \subsubsection{Residual Multi-Scale Block}
The core innovation of our architecture is the Residual Multi-Scale Block (RMSB), which extracts features using parallel convolutional paths:
    \end{multicols}
 \end{minipage}
\newpage
\onecolumn
 \begin{minipage}{\textwidth}
 \setlength{\parindent}{15pt}
    \begin{multicols}{2} %
    \begin{itemize}
   \item $3\times3$ depthwise convolution for local feature extraction.
   \item $5\times5$ depthwise convolution for broader contextual information.
   \item $1\times1$ pointwise convolution for channel transformation.
\end{itemize}
\noindent The outputs of these convolutions are concatenated along the channel dimension and then projected using a $1\times1$ convolution. A residual shortcut connection ensures stable gradient flow during backpropagation, allowing the network to better propagate gradients and train deeper layers. As shown in Figure~\ref{fig:Figure3}, the RMSB operation integrates these features effectively. Formally, for an input $x$  the RMSB operation is defined as:

\[
\begin{aligned}
    \text{RMSB}(x) &= \text{ReLU}\Bigl(\text{BN}\bigl(W_2[\text{DWConv}_{3\times3}(x); \\
    &\quad \text{DWConv}_{5\times5}(x); \text{Conv}_{1\times1}(x)]\bigr) + W_sx\Bigr)
\end{aligned}
\]
   \noindent where $W_s$ is a $1\times1$ convolution when the input and output channel dimensions differ. This design enables the network to capture both local and contextual features efficiently with minimal parameter overhead.
 \subsubsection{Squeeze-and-Excitation Attention}
   To enhance channel-wise feature importance, a Squeeze-and-Excitation (SE) block is applied after the final residual stage. The SE mechanism operates in three steps:
   \begin{enumerate}
    \item \textbf{Squeeze:} Global average pooling compresses spatial information into channel descriptors
    \item \textbf{Excitation:} Two fully connected layers with ReLU and sigmoid activations generate channel attention weights
    \item \textbf{Reweight:} Original features are multiplied by the attention weights
  \end{enumerate}
This mechanism adaptively emphasizes informative features while suppressing less relevant ones, improving discrimination between visually similar disease patterns.
    \subsubsection{Network Architecture}
The overall LWMSCNN-SE architecture consists of seven main components:
\begin{enumerate}
    \item Initial convolution (32 filters, $3\times3$, stride 2) with batch normalization and ReLU.
    \item Three residual multi-scale blocks with increasing channel depths (64, 128, 256).
    \item Depthwise convolutional downsampling layers between blocks (stride 2).
    \item SE attention module with reduction ratio $r=8$.
    \item Global average pooling layer.
    
    \columnbreak
   \item Dense layer with 192 neurons and dropout (0.4).
    \item Softmax classifier with 4 output neurons.
\end{enumerate}
    
\section{Experimental Results}
\subsection{Training Configuration}
The model was trained selected hyperparameters to balance convergence speed, generalization capability, and computational efficiency. Key hyperparameters were determined through empirical validation on the maize disease dataset, with specific values chosen to optimize performance while maintaining training stability. The complete hyperparameter configuration is summarized in Table~\ref{tab:hyperparameters}.

 \begin{table}[H]
     \centering
     \caption{Hyperparameter Configuration for LWMSCNN-SE Training}
     \label{tab:hyperparameters}
     \begin{tabular}{@{}lll@{}}
     \hline
     \textbf{Parameter} & \textbf{Value} & \textbf{Description} \\ \hline
      Image\_size & 224×224 & Input image dimensions \\
     Batch\_size & 32 & Mini-batch size for training \\
     Max\_epochs & 40 & Maximum training epochs \\
     Learning\_rate & 0.0001 & Adam optimizer learning rate \\
     Shuffle\_buffer & 3000 & Dataset shuffle buffer size \\
     Early\_stop\_Patience & 8 & Early stopping patience \\
     Optimizer & Adam & Optimization algorithm \\
     Optimizer\_Params & $\beta_1=0.9, \beta_2=0.999$ & Adam momentum parameters \\  
     Random\_State & 42 & Reproducibility seed \\ \hline
\end{tabular}
\end{table}
\subsection{Classification Report and Confusion Matrix}
The detailed per-class performance metrics are presented in Table~\ref{tab:classification_report}. The model demonstrated exceptional classification capabilities, achieving perfect precision and recall (1.00) for both the Common Rust and Healthy classes. For Gray Leaf Spot, the model maintained strong performance with precision and recall of 0.88, yielding an F1-score of 0.88. Northern Leaf Blight classification also showed excellent results with precision of 0.94 and recall of 0.93, producing an F1-score of 0.93. The overall weighted average across all classes reached 0.97 for precision, recall, and F1-score, confirming the model's balanced and robust classification performance across all disease categories.

The confusion matrix, illustrated in Figure~\ref{fig:Figure4}, provides detailed insight into the model's classification patterns. The matrix reveals minimal misclassifications, with the majority of errors occurring between Gray Leaf Spot and Northern Leaf Blight classes—two conditions that share similar visual char-

    \end{multicols}
 \end{minipage}

\begin{figure}[h!]
    \centering
    \includegraphics[width=0.8\textwidth]{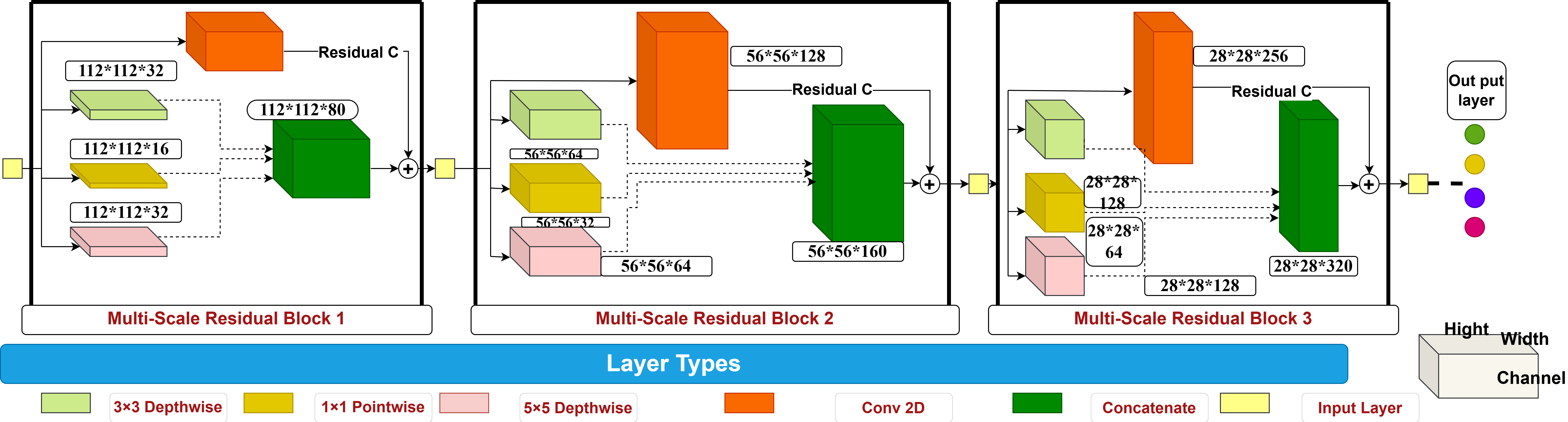}
    \caption{The architecture of the Residual Multi-Scale Block (RMSB).}
    \label{fig:Figure3}
   \end{figure}
\newpage
\twocolumn
\noindent acteristics in maize leaves.This analysis confirms the model's strong discriminatory power while identifying areas for potential improvement in distinguishing between visually similar disease manifestations.

\begin{table}[H]
\centering
\caption{Classification Report on Test Set (386 samples)}
\label{tab:classification_report}
\begin{tabular}{@{}lcccc@{}}
\hline
\textbf{Class} & \textbf{Precision} & \textbf{Recall} & \textbf{F1-Score} & \textbf{Support} \\  \hline
Gray Leaf Spot & 0.88 & 0.88 & 0.88 & 51 \\
Common Rust & 1.00 & 1.00 & 1.00 & 120 \\
Northern Leaf Blight & 0.94 & 0.93 & 0.93 & 99 \\
Healthy & 0.99 & 1.00 & 1.00 & 116 \\  \hline
\textbf{Accuracy} & & & \textbf{0.97} & \textbf{386} \\
\textbf{Macro Average} & 0.95 & 0.95 & 0.95 & 386 \\
\textbf{Weighted Average} & 0.97 & 0.97 & 0.97 & 386 \\  \hline
\end{tabular}
\end{table}
\noindent
\begin{figure}[H]
    \centering
    \includegraphics[width=0.5\textwidth]{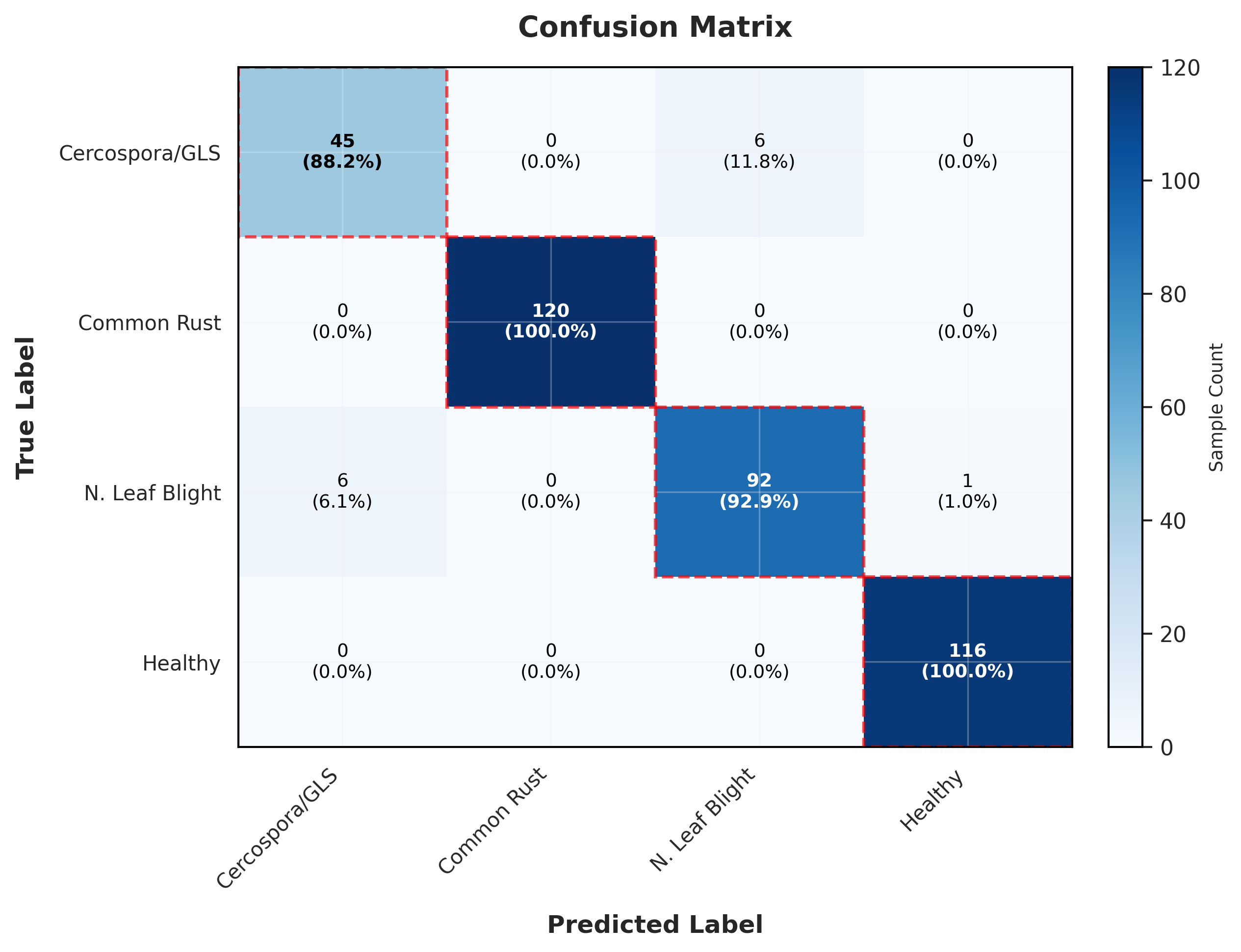} 
    \caption{Confusion Matrix for Maize Disease Classification.}
    \label{fig:Figure4}
    \end{figure}
 \subsection{Training Performance Analysis and Visualization}
\begin{figure}[H]
\centering

\includegraphics[width=1\textwidth]{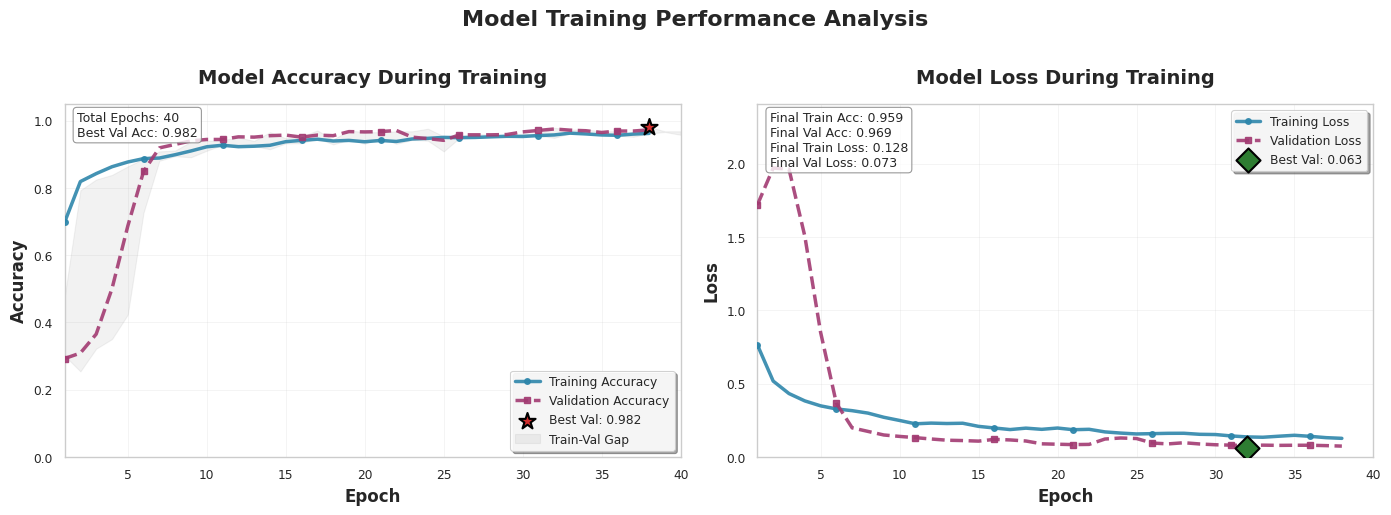}
\caption{Model training performance: Accuracy and loss curves.}
\label{fig:Figure5}
\end{figure}

     The training dynamics of the proposed LWMSCNN-SE model are comprehensively analyzed through the visualization presented in Figures~\ref{fig:Figure5}. As shown in the training curves, the model demonstrates strong convergence behavior with training accuracy reaching 95.85\% and validation accuracy achieving 96.88\% after 40 epochs. The training loss decreases steadily from 1.1714 to 0.1283, while validation loss shows similar improvement from 1.4339 to 0.0726, indicating effective learning without significant overfitting. The validation accuracy stabilizes above 96\% after epoch 17, maintaining this high performance level throughout the remaining training cycles. These visualizations collectively demonstrate the model's robust learning capacity, stable convergence behavior, and excellent generalization characteristics, validating the effectiveness of the proposed architectural design and training strategy for maize disease classification tasks. 
    \subsection{Comparison with Existing Models}
     Table \ref{table:model_comparison} shows the comparison between LWMSCNN-SE and conventional CNN architectures such as ResNet50, VGG16, EfficientNet-B0, and MobileNetV1. In all models, all layers are fine-tuned using pretrained weights from ImageNet to adapt the models to the target task. The custom head added to each model includes a Global Average Pooling layer, followed by a Dense layer (192 units, ReLU activation), a Dropout layer(0.4 rate), and a final Dense output layer with softmax activation for multi-class classification. Compared to ResNet50, LWMSCNN-SE reduces parameters by 99.0\% and GFLOPs by 91.4\% while maintaining comparable accuracy. Compared to VGG16, LWMSCNN-SE reduces parameters by 98.37\% and GFLOPs by 97.83\%, with similar accuracy. In comparison to EfficientNet-B0 and MobileNetV1, LWMSCNN-SE maintains competitive accuracy with a significant reduction in computational cost, making it highly efficient for deployment in resource-constrained environments. This makes LWMSCNN-SE a highly effective and lightweight alternative to traditional CNN architectures.
\begin{table}[H]
 \centering
\caption{Comparison of LWMSCNN-SE with conventional CNN architectures in terms of parameters, GFLOPs, and accuracy.}
\label{table:model_comparison}
\begin{tabular}{@{}lcccc@{}}
\hline
\textbf{Model}          & \textbf{Parameters}   & \textbf{GFLOPs} & \textbf{Accuracy} & \textbf{Efficiency}     \\ \hline
ResNet50               & 23,981,892           & 7.752           & 97.67\%           & 12.61\%               \\
VGG16                  & 14,813,956           & 30.713          & 97.15\%           & 3.16\%                \\
EfficientNet-B0        & 4,254,272            & 0.801           & 94.3\%            & 117.83\%              \\
MobileNetV1            & 3,404,548            & 1.146           & 96.89\%           & 84.72\%               \\
\textbf{LWMSCNN-SE}    & \textbf{241,348}     & \textbf{0.666}  & \textbf{96.63\%}  & \textbf{145.31\%}     \\ \hline
\end{tabular}
\end{table}
\subsection{Ablation Study}
An ablation study was conducted to evaluate the contribution of individual components in the LWMSCNN-SE architecture. As summarized in Table~\ref{table:ablution}, the full model configuration achieved the highest accuracy of 96.98\% with 241,348 parameters and 0.666 GFLOPS. Removing the Squeeze-and-Excitation attention mechanism resulted in a noticeable accuracy drop to 94.81\% while reducing parameters to 224,968 and GFLOPS to 0.665, demonstrating the importance of channel-wise attention for feature recalibration. Eliminating data augmentation during training reduced accuracy to 95.077\% while maintaining the same parameter count and computational complexity, highlighting the critical role of augmentation in preventing overfitting and improving generalization. Replacing the multi-scale depthwise convolutions with standard convolutional layers significantly increased both parameters (640,100) and computational cost (1.876 GFLOPS) while reducing accuracy to 94.30\%, confirming the efficiency benefits of depthwise separable convolutions. Finally, removing residual connections while maintaining downsampling layers resulted in 456,508 parameters and 1.419 GFLOPS with substantially reduced performance, underscoring the importance of residual learning for gradient flow and feature reuse. These results collectively validate the design choices in the proposed LWMSCNN-SE architecture.
     
\section{Conclusion}

This paper presented LWMSCNN-SE, a lightweight convolutional neural network designed for efficient and accurate maize disease classification in resource-constrained environments. By innovatively integrating residual multi-scale depthwise blocks with a Squeeze-and-Excitation channel attention mechanism, the model achieves a high classification accuracy of 96.63\% 
     
\newpage
\onecolumn

\begin{table}[ht]
\centering
\caption{Ablation Study Results for LWMSCNN-SE Components}
\label{table:ablution}
\begin{tabular}{lllll}
\hline
\textbf{Configuration}                     & \textbf{Accuracy (\%)} & \textbf{Parameters} & \textbf{GFLOPS} & \textbf{Relative to Full Model} \\ \hline
\textbf{Full LWMSCNN-SE}                            & \textbf{96.63 }         & \textbf{241,348}           & \textbf{0.666}         & \textbf{Reference}                        \\ 
Without SE Block                          & 94.81                  & 224,968             & 0.665           & $\downarrow$2.17\% accuracy       \\ 
Without Data Augmentation                  & 95.077                 & 241,348             & 0.666           & $\downarrow$1.903\% accuracy     \\ 
Without Depthwise Convolutions             & 94.30                  & 640,100             & 1.876           & $\uparrow$165\% parameters, $\uparrow$182\% GFLOPS \\ 
Without Residual Blocks                    & 95.595                 & 456,508             & 1.419           & $\uparrow$89\% parameters, $\uparrow$113\% GFLOPS \\ \hline
\end{tabular}
\end{table}

\begin{minipage}{\textwidth}
 \setlength{\parindent}{15pt}
    \begin{multicols}{2} %
\noindent while maintaining an exceptionally low computational footprint of only 241,348 parameters and 0.666 GFLOPs.

The proposed architecture effectively addresses the critical accuracy-efficiency trade-off prevalent in deploying deep learning models for precision agriculture. The multi-scale blocks capture discriminative features at various receptive fields, essential for identifying disease patterns of different sizes and morphologies. Simultaneously, the SE block dynamically recalibrates channel-wise feature responses, enhancing the model's ability to distinguish between visually similar diseases. As demonstrated through comprehensive comparisons and an ablation study, this synergistic design outperforms or matches the accuracy of substantially larger models like VGG16 and ResNet50 while being orders of magnitude more efficient than general-purpose lightweight networks like MobileNetV1.

The results confirm that LWMSCNN-SE is highly suitable for real-time, in-field deployment on edge devices such as smartphones, drones, and IoT sensors, offering a practical tool for early disease diagnosis and management. Future work will focus on hardware-aware optimizations, including quantization and pruning for further compression, deployment and latency testing on embedded platforms (e.g., Raspberry Pi, NVIDIA Jetson Nano), and expanding the model's robustness by training on larger, more diverse datasets captured under challenging field conditions.

\section*{Acknowledgment}
The authors would like to express their deepest gratitude to **Beri** and **Alas**, whose unwavering support and strength have been a guiding force throughout this journey. Their memory continues to inspire and empower this work. 
   
     \columnbreak

\vspace{0.5cm}

\section*{Data Availability}
The PlantVillage dataset is publicly available on Kaggle:\\
\url{https://www.kaggle.com/datasets/abdallahalidev/plantvillage-dataset}

 \end{multicols}
\end{minipage}

\newpage
\twocolumn


\begin{thebibliography}{00}
\bibitem{Sibiya2019}
M. Sibiya and M. Sumbwanyambe, "A computational procedure for the recognition and classification of maize leaf diseases out of healthy leaves using convolutional neural networks," *IEEE/CVF Conference on Computer Vision and Pattern Recognition*, 2019, pp. 147–160.

\bibitem{subbarayudu2025}
C. Subbarayudu and M. Kubendiran, "Segmentation-based lightweight multi-class classification model for crop disease detection, classification, and severity assessment using DCNN," *IEEE Conference on Artificial Intelligence in Agriculture*, 2025, pp. 1–10.
\bibitem{nsibo2024}
D. Nsibo, I. Barnes, and D. Berger, "Recent advances in the population biology and management of maize foliar fungal pathogens *Exserohilum turcicum*, *Cercospora zeina*, and *Bipolaris maydis* in Africa," *IEEE Conference on Plant Science and Technology*, 2024, Art. no. 1404483.


\bibitem{savary2019}
S. Savary, L. Willocquet, S. J. Pethybridge, P. D. Esker, N. McRoberts, and A. Nelson, "The global burden of pathogens and pests on major food crops," *IEEE International Conference on Agricultural Sciences*, 2019, pp. 430–439.

\bibitem{sharma2025}
N. Sharma, P. Sharma, and N. Kumar, "Feature engineering to early detection of plant disease using image processing and artificial intelligence: A comparative analysis," *IEEE International Conference on Artificial Intelligence in Agriculture*, 2025, pp. 100–110.

\bibitem{dulla2025}
J. Dulla, "Plant disease detection using deep learning," *IEEE Conference on Deep Learning in Agriculture*, 2025, pp. 50–60.

\bibitem{sladojevic2016}
S. Sladojevic, M. Arsenovic, A. Anderla, D. Ćulibrk, and D. Stefanović, "Deep neural networks based recognition of plant diseases by leaf image classification," *IEEE International Conference on Computational Intelligence and Neuroscience*, 2016, Art. no. 3289801.

\bibitem{lu2021}
J. Lu, L. Tan, and H. Jiang, "Review on convolutional neural network (CNN) applied to plant leaf disease classification," *IEEE Conference on Agricultural Engineering and Technology*, 2021, pp. 707–715.

\bibitem{abade2020}
A. Abade, P. Ferreira, and F. Vidal, "Plant diseases recognition on images using convolutional neural networks: A systematic review," *IEEE International Conference on Computer Vision in Agriculture*, 2020, pp. 106–125.

\bibitem{shafik2024}
W. Shafik, A. Tufail, C. De Silva Liyanage, and R. Apong, "Using transfer learning-based plant disease classification and detection for sustainable agriculture," *IEEE Conference on Smart Agriculture and Sustainable Practices*, 2024, pp. 85–92.

\bibitem{dixit2025}
P. Dixit, C. Andhale, and S. Jaiswar, "Fruit disease detection using VGG16 and Flask: A deep learning-based web application for precision agriculture," *IEEE Conference on Precision Agriculture and Technology*, 2025, pp. 50–60.


\bibitem{khan2023}
A. Khan, S. Jensen, A. Khan, and S. Li, "Plant disease detection model for edge computing devices," *IEEE International Conference on Edge Computing for Agriculture*, 2023, pp. 1025–1030.



\bibitem{jiang2022}
J. Jiang, H. Liu, C. Zhao, C. He, J. Cheng, Y. Zhu, W. Cao, and X. Yao, "Evaluation of diverse convolutional neural networks and training strategies for wheat leaf disease identification with field-acquired photographs," *IEEE International Conference on Remote Sensing in Agriculture*, 2022, pp. 3446–3450.

\bibitem{xiang2021}
S. Xiang, Q. Liang, W. Sun, D. Zhang, and Y. Wang, "L-CSMS: novel lightweight network for plant disease severity recognition," *IEEE International Conference on Plant Disease Management*, 2021, pp. 557–569.

\bibitem{Haque2022}
M. Haque, S. Marwaha, C. Deb, S. Nigam, A. Arora, K. Hooda, P. Soujanya, S. Aggarwal, B. Lall, M. Kumar, S. Islam, M. Panwar, P. Kumar, and R. Agrawal, "Deep learning-based approach for identification of diseases of maize crop," *Scientific Reports*, vol. 12, 2022, Art. no. 10140.
\bibitem{Joseph2024}
D. Joseph, P. Pawar, and K. Chakradeo, "Real-Time Plant Disease Dataset Development and Detection of Plant Disease Using Deep Learning," *IEEE Access*, vol. 12, pp. 16310–16333, 2024.

\bibitem{Shi2023}
T. Shi, Y. Liu, X. Zheng, K. Hu, H. Huang, H. Liu, and H. Huang, "Recent advances in plant disease severity assessment using convolutional neural networks," *Scientific Reports*, vol. 13, 2023, Art. no. 29230.

\bibitem{FKhan2023}
F. Khan, N. Zafar, M. Tahir, M. Aqib, H. Waheed, and Z. Haroon, "A mobile-based system for maize plant leaf disease detection and classification using deep learning," *Frontiers in Plant Science*, vol. 14, 2023, Art. no. 1079366.


\bibitem{Andrew2022}
A. J., J. Eunice, D. E. Popescu, M. K. Chowdary, and J. Hemanth, "Deep learning-based leaf disease detection in crops using images for agricultural applications," *Agronomy*, vol. 12, no. 10, p. 2395, 2022.

\bibitem{Chowdhury2021}
M. E. H. Chowdhury, T. Rahman, A. Khandakar, M. A. Ayari, A. U. Khan, M. S. Khan, N. Al-Emadi, M. B. I. Reaz, M. T. Islam, and S. H. Md Ali, "Automatic and reliable leaf disease detection using deep learning techniques," *AgriEngineering*, vol. 3, no. 2, pp. 294–312, 2021.

\bibitem{Li2024}
R. Li, Y. Li, W. Qin, A. Abbas, S. Li, R. Ji, Y. Wu, Y. He, and J. Yang, "Lightweight network for corn leaf disease identification based on improved YOLO v8s," *Agriculture*, vol. 14, no. 2, p. 220, 2024.

\bibitem{Li2023}
G. Li, Y. Wang, Q. Zhao, P. Yuan, and B. Chang, "PMVT: A lightweight vision transformer for plant disease identification on mobile devices," *Frontiers in Plant Science*, vol. 14, 2023, Art. no. 1256773.

\bibitem{Howard2017}
A. G. Howard et al., "MobileNets: Efficient convolutional neural networks for mobile vision applications," *arXiv:1704.04861*, 2017.

\bibitem{Zhang2018}
X. Zhang, X. Zhou, and M. Lin, "ShuffleNet: An extremely efficient convolutional neural network for mobile devices," *2018 IEEE/CVF Conference on Computer Vision and Pattern Recognition*, pp. 6848-6856, 2018.


\bibitem{Tan2019}
M. Tan and Q. Le, "EfficientNet: Rethinking model scaling for convolutional neural networks," *ICML*, 2019.

\bibitem{Nguyen2020}
H. Nguyen, "A lightweight and efficient deep convolutional neural network based on depthwise dilated separable convolution," 2020.

\bibitem{Wang2025}
X. Wang, "EfficientNet-SE: A lightweight attention-enhanced CNN for real-time crop leaf disease detection in precision agriculture," in *2025 6th International Conference on Internet of Things, Artificial Intelligence and Mechanical Automation (IoTAIMA)*, 2025, pp. 63–67.

\bibitem{AlGaashani2025}
M. S. A. M. Al-Gaashani, R. Alkanhel, M. A. S. Ali, M. S. A. Muthanna, A. Aziz, and A. Muthanna, "MSCPNet: A multi-scale convolutional pooling network for maize disease classification," *IEEE Access*, vol. 13, pp. 11423–11446, 2025.

\bibitem{Shandilya2025}
G. Shandilya, S. Gupta, H. G. Mohamed, S. Bharany, A. U. Rehman, and S. Hussen, "Enhanced maize leaf disease detection and classification using an integrated CNN-ViT model," *Food Science and Nutrition*, vol. 13, no. 7, p. e70513, 2025.


\end{thebibliography}
\end{document}